\documentclass[conference]{IEEEtran}
\usepackage{amsfonts}
\usepackage{graphicx}
\usepackage{cite}

\ifCLASSINFOpdf
\else
\fi
\begin{document}
%
\title{Quantification of observed prior and likelihood information in parametric Bayesian modeling}

\author{\IEEEauthorblockN{Giri Gopalan}
\IEEEauthorblockA{Doctoral Student in the School of Engineering and Natural Sciences; University of Iceland\\
}}


%


\maketitle

\begin{abstract}
Two data-dependent information metrics are developed to quantify the information of the prior and likelihood functions within a parametric Bayesian model, one of which is closely related to the reference priors from \cite{berger2009} and information measure introduced in \cite{lindley1956}. A combination of theoretical, empirical, and computational support provides evidence that these information-theoretic metrics may be useful diagnostic tools when performing a Bayesian analysis.
\end{abstract}


%
\IEEEpeerreviewmaketitle

\section{Introduction}
\subsection{Introduction}
Consider a general problem of statistical inference where unknowns are represented by $\theta$ and knowns are represented by $Y_{obs}$; the Bayesian approach is to solve, sample from, or approximate the posterior distribution $p(\theta|Y_{obs})$ and associated quantities. The posterior distribution is proportional to the product of the prior function (the marginal probability model for the unknowns) and  the likelihood function (a probability model for the observed data given the unknowns). Since these functions are the key assumptions embedded into a Bayesian model, it is natural to quantify their strength. In other words, just how much does $Y_{obs}$, the data collected in an experiment, influence the inference of $\theta$ in comparison to a prior function?  The overarching objective of this paper is to define and critically examine two data-dependent metrics\footnote{To the potential chagrin of some, this paper uses the terms metric and measure synonymously. Additionally, the word metric is not used in the real-analytic sense, and the KL-divergence is not a metric according to the real-analytic definition due to violation of symmetry and the triangle inequality.}  that attempt to answer this question. 

The Bayesian viewpoint has been criticized due to the challenge of appropriately selecting a prior distribution, and a variety of approaches have been taken to construct a default prior or bypass the prior completely, such as the Jeffreys prior \cite{1946RSPSA.186..453J}, reference priors \cite{berger2009}, prior free inference \cite{martin2013inferential}, and other approaches reviewed in \cite{10.2307/2291752}. Furthermore, in \cite{yuan1999minimally}, information theoretic arguments are introduced to determine a default likelihood. In contrast, this work does not attempt to introduce a default prior, likelihood, nor a fundamentally new inferential procedure. Instead, it suggests using standard parametric Bayesian inference with an arbitrarily specified prior and likelihood (perhaps one of the well studied default choices), and a pair of data-dependent information-theoretic metrics to quantify the information of both the prior and likelihood functions chosen. While this approach does not provide an explicit set of rules to define a prior or likelihood, it allows one to use a quantity to determine if prior or data assumptions are too strong or weak, similarly to how an analyst might use a p-value as a measure or tolerance of extremity under an assumed probabilistic model. 

\subsection{Definition of Prior and Likelihood Information}
Before setting forth the definitions of prior and likelihood information, it is important to clarify the notation that will be used in this work.  The notational conventions from \cite{gelman2013bayesian} are adopted unless otherwise stated. Additionally, $D_{KL}(.,.)$ refers to the Kullback-Leibler divergence (KL-divergence; \cite{information_theory_statistics}) from the first random variable (or probability density, to slightly abuse notation) to the second random variable (or probability density); that is, the expectation in the definition of the KL-divergence is with respect to the first random variable's distribution. For instance, $D_{KL}(p(x),q(x))$ for continuous random variables $P$ and $Q$, whose densities are $p(x)$ and $q(x)$, is:
\begin{eqnarray*}
D_{KL}(p(x),q(x)) &=& D_{KL}(P,Q) \\
&=& \int Log[\frac{p(x)}{q(x)}]p(x)dx
\end{eqnarray*}
Let $\theta$ be a particular parameter(s) of interest.
\subsubsection{Definition 1.1; Normalized likelihood} The normalized likelihood $L_{\theta}(\theta)$ is defined as $\frac{p(y_{obs}|\theta)}{\int p(y_{obs}|\theta) d\theta}$. By definition, this is dependent on the particular parameterization, $\theta$, that is chosen for the analysis (emphasized by the subscript $L_{\theta}(.)$).
\subsubsection{Definition 1.2; Prior information} $u = D_{KL}(p(\theta|y_{obs}),L_{\theta}(\theta))$.
\subsubsection{Definition 1.3; Likelihood information} $v = D_{KL}(p(\theta|y_{obs}),p(\theta))$.\\
Conceptually, the information of the data (via the likelihood function) is judged by the distance from the posterior to the prior relative to the posterior; likewise, the information of the prior is the distance from the posterior to the normalized likelihood relative to the posterior. The influence of the prior distribution is thus quantified by the discrepancy between the posterior distribution and likelihood \footnote{It is possible that the likelihood is not integrable, limiting the applicability of prior information. However, in many cases likelihood integrability may be achieved by assuming a compact parameter space a priori, which is often a scientifically plausible assumption.}, which is often used if a prior distribution is absent  (i.e., in a likelihood based method of inference). 

\subsection{A Motivating Example Regarding Prior and Likelihood Information}
Here a scenario is developed where a ``noninformative" Jeffreys  prior and ``noninformative" reference prior are more informative than a flat prior. Consider data that is collected according to a bivariate binomial model, whose probability mass function is given by:
\begin{eqnarray*}
f(r,s|p,q,m) &=& {m \choose r}p^{r}(1-p)^{m-r}{r \choose s}q^{s}(1-q)^{r-s}
\end{eqnarray*}
The observed data are $r$ and $s$, and the inferential parameters of interest are $p$ and $q$.  The reference prior as in \cite{yang} is:
\begin{eqnarray*}
\pi_{ref}(p,q) &=& (\pi)^{-2}p^{-1/2}(1 - p)^{-1/2}q^{-1/2}(1 - q)^{-1/2}
\end{eqnarray*}
The Jeffreys prior is:
\begin{eqnarray*}
\pi_{Jeff}(p,q) &=& (2\pi)^{-1}(1-p)^{-1/2}q^{-1/2}(1-q)^{-1/2}
\end{eqnarray*}
Note the distinction between subscripted $\pi$ to refer to a function and $\pi$ the numerical constant; also the $p(.)$ notation is avoided to reduce confusion with the parameter $p$.

Assume that the following data are collected in an experiment: m = 30,  r = 29, and s = 2. Most of the likelihood's mass occupies one of the four corners of the unit square, where both priors approach $\infty$. Hence the reference prior will have more information than a flat prior, and, intuitively, the likelihood will have more information when the prior is flat (assuming that prior and likelihood information are inversely related). Indeed, the numerically computed likelihood information with a flat prior is 3.53, 2.97 for the reference prior, and 2.55 for the Jeffreys prior.\footnote{As defined, the prior information is 0 for a flat prior and strictly positive for the other two priors. To cite this without computing the likelihood information, however, would seem to be tautological.}
How is this apparent inconsistency resolved?  While there exists utility for default priors, this example illustrates it is important to quantify observed prior and likelihood information.

The structure of the paper is as follows: in section 2, some theoretical properties of the prior and likelihood information metrics are reviewed and developed, in section 3, it is shown that the information metrics can be computed analytically in some basic conjugate models, and in section 4 these metrics are applied to a few prediction problems with non-conjugate priors, illustrating that small prior information may be beneficial for the predictive accuracy of a Bayesian model. 

\section{Important Theoretical Properties of Prior and Likelihood Information}
This section reviews and develops some important theoretical properties of prior and likelihood information; namely, these properties are the invariance property of the likelihood information, the interpretation of the likelihood information as observed mutual information between $Y_{obs}$ and $\theta$, and the decay of prior information to 0 as more data is collected in commonly used Bayesian models.  Additionally, an important motivation for the use of the KL-divergence in defining prior and likelihood information is that it is non-negative and 0 if and only if the probability measures compared are identical almost surely \cite{Cover:2006:EIT:1146355}. However, the second section of the appendix considers when the prior or likelihood are not necessarily integrable, in which case it is still possible that the likelihood information metric is defined, but the guarantee of non-negativity is lost.

A key property of likelihood information is that it is invariant to 1-1 reparameterization due to properties of the KL-divergence. However, the same property does not hold for prior information since a Jacobian is not necessary for the normalized likelihood when starting with a different parameterization; in other words $L_{\phi}(\phi)$ is proportional to $L_{\theta}(\theta(\phi))$. Results on the invariance property of the KL-divergence can be found, for example, in \cite{information_theory_statistics}. Hence, under the proposed definition of likelihood information, an analyst can be assured that the amount of information in the data observed in an experiment does not change with respect to different model parameterizations. A measure of data informativeness which changes depending on the mathematical specification of the model used is not desirable from a scientific point of view.

Typically, the reference prior maximizes average likelihood information, which is equivalent to maximizing the mutual information between $Y_{obs}$ and $\theta$. 
From this perspective, the likelihood information can be considered the observed mutual information between $Y_{obs}$ and $\theta$. Therefore, the relationship between the likelihood information and the mutual information between $Y_{obs}$ and $\theta$ is analogous to the relationship between the observed and expected Fisher information, where the critical distinction between these quantities is that the expected Fisher information is an average over data. Furthermore, this property connects the information measure in Definition 1 of \cite{lindley1956} with the likelihood information metric, since both measures yield the mutual information between $Y_{obs}$ and $\theta$ when averaged over $Y_{obs}$. Moreover, this demonstrates that observed mutual information is not unique. However, in contrast to the information measure in Definition 1 of \cite{lindley1956}, the likelihood information is invariant to 1-1 reparameterization without taking an average over $Y_{obs}$.  It should be stressed that the information metrics developed within this paper are not averaged over data, but are instead functionally dependent on data. Therefore, it is appropriate to refer to them as observed information metrics as opposed to expected ones. 

Additionally, it should be noted that Lehmann and Casella \cite{lehmann} suggest that the KL-divergence from the posterior to the prior is the information ``between the data and the parameter". The perspective taken in this paper is that this is most aptly characterized as the information provided by the likelihood, and, hence, the observed data.

What follows is a proof that the prior information goes to zero in probability when the parameter space is finite, the model is correctly specified, and data are generated i.i.d from this model.

\textbf{Theorem:}
Assume the parameter space $\Theta$ is a finite set which contains the true parameter $\theta_0$, $p(y_{obs}|\theta) > 0$, $p(\theta) > 0$, $\forall \theta \in \Theta$ and $\forall y_{obs}$, and data are generated i.i.d according to a true model $p(y_{obs}|\theta_0)$, where the model is correctly specified. Under these asssumptions, the prior information approaches 0 in probability.

\textbf{Proof:} By posterior consistency, as in Appendix B of \cite{gelman2013bayesian}, the probability masses on $\theta$ governed by the posterior and normalized likelihood (which is a posterior under a flat prior) both converge in probability to mass of 1 on $\theta_0$ and 0 elsewhere. The KL-divergence between the posterior and normalized likelihood is a continuous map that is a function of the probability masses specified by the posterior and normalized likelihood. Therefore, the continuous mapping theorem applies. In particular, the sequence of KL-divergences between the posterior and likelihood converges in probability to the KL-divergence between two point masses at $\theta_0$, which is 0.

Additionally, a conjecture in the continuous case is as follows: due to posterior consistency, both the log-likelihood and log-posterior can be reasonably approximated with a Taylor expansion about the truth, so both $L_{\theta}(\theta)$ and $p(\theta|y_{obs})$ can be approximated as a normal distribution with $\theta_0$ as the mean and the inverse of the Fisher information at $\theta_0$ as the variance (at least when the observed data are generated i.i.d conditioned on the underlying parameters, and sufficient regularity conditions are met as in the Bernstein von Mises theorem \cite{asymptotics}). Therefore, the KL-divergence between $L_{\theta}(\theta)$ and $p(\theta |y_{obs})$ approaches 0; techniques used  in \cite{746784} might be useful for proving this claim.

\section{Prior and Likelihood Information in Basic Conjugate Models}
To illustrate that prior and likelihood information can be used in practice, closed form expressions are computed for the prior and likelihood information in the normal-normal and multinomial-Dirichlet models. These results may be useful in deriving the prior and likelihood information in models that use conjugate priors, such as `naive Bayes classification', which can be thought of as a pair of multinomial-Dirichlet models for each of two classes. More generally, the third part of the appendix presents a result which demonstrates why the normalized likelihood, and hence prior information, may be defined when the likelihood model is assumed to come from the exponential family. 

To clarify notation, a random variable denoted by $\theta*$ is distributed according to the normalized likelihood, $L_{\theta}(\theta)$. 

\subsection{Normal-Normal Model With Known Variance}
Assume $n$ i.i.d samples $y_i|\mu \sim N(\mu, \sigma^2)$ where the variance $\sigma^2$ is known and the mean parameter $\mu \sim N(\mu_0,\sigma_0^2)$. The KL - divergence between the posterior and prior can be computed by making use of the fact that the KL - divergence from $N_1 \sim N(\mu_1,\sigma_1^2)$ to $N_2 \sim N(\mu_2,\sigma_2^2)$ is given by
\begin{eqnarray*}
D_{KL}(N_1,N_2) &=& \frac{(\mu_1-\mu_2)^2+\sigma_1^2-\sigma_2^2}{2\sigma_2^2} + Log(\frac{\sigma_2}{\sigma_1})
\end{eqnarray*}
as in \cite{KL}. In the normal-normal model, the posterior is given by:
\begin{eqnarray*}
\mu | y \sim N(\frac{\frac{\mu_0}{\sigma_0^2}+\frac{n\bar{y}}{\sigma^2}}{\frac{1}{\sigma_0^2}+\frac{n}{\sigma^2}},\frac{1}{\frac{1}{\sigma_0^2}+\frac{n}{\sigma^2}})
\end{eqnarray*}
Hence, substituting the latter equation into the former and simplifying, $D_{KL}(\mu|\bar{y},\mu)  = [(\mu_0-\frac{\sigma^2\mu_0+n\bar{y}\sigma_0^2}{\sigma^2+n\sigma_0^2})^2- \sigma_0^2 + \frac{1}{\frac{1}{\sigma_0^2}+\frac{n}{\sigma^2}}]\frac{1}{2\sigma_0^2} + Log(\sigma_0\sqrt{\frac{1}{\sigma_0^2}+\frac{n}{\sigma^2}})$
which yields the KL-divergence between the posterior and prior, or likelihood information. To calculate the prior information, note that:
\begin{eqnarray*}
\mu* \sim N(\bar{y},\sigma^2/n)
\end{eqnarray*}
Hence the prior information is:
$ D_{KL}(\mu|\bar{y},\mu*) = [(\bar{y}-\frac{\sigma^2\mu_0+n\bar{y}\sigma_0^2}{\sigma^2+n\sigma_0^2})^2-\frac{\sigma^2}{n} + \frac{1}{\frac{1}{\sigma_0^2}+\frac{n}{\sigma^2}}]\frac{n}{2\sigma^2}+ Log(\frac{\sigma\sqrt{\frac{1}{\sigma_0^2}+\frac{n}{\sigma^2}}}{\sqrt{n}})$.
In the  normal-normal model one can use the law of iterated expectation to show that:
\begin{eqnarray*}
E[\bar{y}] = \mu_0
\end{eqnarray*}
and 
\begin{eqnarray*}
E[\bar{y}^2] = \sigma_0^2+\mu_0^2+\sigma^2/n
\end{eqnarray*}
If $\sigma=\sigma_0 = 1$ and $\mu_0 = 0$ (for computational convenience), the average prior information is:
\begin{eqnarray*}
E_{Y}[D_{KL}(\mu|\bar{y},\mu*)] &=& Log[\sqrt{\frac{n+1}{n}}]
\end{eqnarray*}
Since the limit of this is 0 as $n$ approaches $\infty$, by Markov's inequality the prior information approaches 0 in probability. It is interesting that in the case when the data are generated according to the marginal distribution of $Y$, the posterior does not always contract to a fixed point, despite that the prior information approaches 0 in probability. This is because the variance of $\bar{y}$ is 1 as $n$ approaches $\infty$. 

\subsection{Multinomial-Dirichlet Model}
Assume a vector $x \in \mathbb{R}^K$ is drawn from a multinomial distribution with probabilities $(p_1,..p_K)$, and those probabilities are drawn from a Dirichlet distribution with concentration parameters $(\alpha_1,..\alpha_K)$. To derive the prior and likelihood information, note that the KL-divergence between $D_1 \sim Dirichlet(\alpha)$ and $D_2 \sim Dirichlet(\beta)$ is given by: $
D_{KL}(D_1,D_2) = Log\Gamma(\alpha_0)-\sum_{i=1}^K Log\Gamma(\alpha_i)-Log\Gamma(\beta_{0})+\sum_{i=1}^KLog\Gamma(\beta_i)+\sum_{i=1}^K(\alpha_{i}-\beta_{i})(\psi(\alpha_i)-\psi(\alpha_0))$
where $\alpha_0 = \sum_{i=1}^K\alpha_i$ and $\beta_0 = \sum_{i=1}^K\beta_i$ \cite{KL}. Also, $p|x \sim Dirichlet(\alpha+x)$ and $p \sim Dirichlet(\alpha)$. Hence the likelihood information $D_{KL}(p|x, p)$ is:
$Log\Gamma(\sum_{i=1}^K\alpha_i + n)-\sum_{i=1}^K Log\Gamma(\alpha_i+x_i)-Log\Gamma(\alpha_0)+\sum_{i=1}^KLog\Gamma(\alpha_i)+\sum_{i=1}^K x_i (\psi(\alpha_i+x_i)-\psi(\sum_{i=1}^K \alpha_i +n))$.
Noting that $p* \sim Dirichlet(x+1)$, the prior information $D_{KL}(p|x,p*)$ is:
$Log\Gamma(\sum_{i=1}^K\alpha_i + n)-\sum_{i=1}^K Log\Gamma(\alpha_i+x_i)- Log\Gamma(n+K)+\sum_{i=1}^KLog\Gamma(x_i+1)+\sum_{i=1}^K(\alpha_i-1) (\psi(\alpha_i+x_i)-\psi(\sum_{i=1}^K \alpha_i +n))$,
where $n = \sum_{i=1}^K x_i$.

\section{Experiments Investigating the Relationship Between Predictive Accuracy and Prior Information}
This section presents and discusses the results of two prediction experiments with Bayesian models, which attempt to understand the relationship between prior information and predictive accuracy; intuitively, as suggested in \cite{gelman2008}, ``weak information" (or regularization) of the prior ought to improve the out of sample classification error of a model. The regularized regression estimates that have been used in recent decades with good predictive performance, such as Lasso \cite{tibshirani1996regression}, can be seen as a departure from typical linear regression estimates that maximize a likelihood. Hence prior information, which quantifies a discrepancy between the posterior and likelihood, ought to be reflected in prediction accuracy.

The first experiment makes use of a machine learning diabetes classification dataset \cite{diabetes} with a logistic regression model. This model is trained using independent normal priors on the coefficients, with varying standard deviations, hence varying the prior and likelihood information content. The dataset consists of two labels (diabetes or no diabetes), 8 continuous predictors, and 758 data points, of which 500 are randomly chosen for training and the remaining are chosen for the test set. Let $Y_{test} \in \{0,1\}^{258}$ and $Y_{train} \in \{0,1\}^{500}$ correspond to the test and training labels for diabetes outcome, respectively.

Let $X_i \in \mathbb{R}^9$ represent the background covariates (and 1 for an offset term) for the $i_{th}$ individual and $\beta \in \mathbb{R}^9$ be the vector of model coefficients. Then, the model used is:
\begin{itemize}
\item $\beta \sim MVN(0,\sigma^2I)$ The variance parameter $\sigma^2$ is fixed, but can be toggled to control the prior and likelihood information.
\item $Y_i|\beta, X_i \sim Bernoulli(logit^{-1}[X_i^{T}\beta])$
\item $Y_i$ are independent conditional on $\beta$
\end{itemize}
For all $i \in 1...758$. 

Then by the definition of conditional probability and marginalization, the posterior predictive distribution for $Y_{test}$ (conditional on $Y_{train}$) is given by $\int_{\mathbb{R}^9} p(Y_{test}|\beta,Y_{train})p(\beta|Y_{train}) d\beta$. For a fixed value of $\sigma^2$, 100 samples were generated from $p(\beta|Y_{train})$, the posterior distribution of model coefficients, using the elliptical slice sampling algorithm \cite{murray2010elliptical}.  These samples were used to draw from the posterior predictive distribution of the diabetes label for the remaining 258 individuals in the study, with the logistic link function and Bernoulli random draws. Retaining the samples for $Y_{test}$ thus generates samples from the posterior prediction of $Y_{test}$ given $Y_{train}$, as discussed in \cite{gelman2013bayesian}. The posterior predictive mode is used for the predictive classifications of the remaining units, which is a prediction justifiable from a decision theoretic viewpoint for a 0-1 loss function. Average 0-1 loss is used as the measure of predictive error, and the results of the experiment are shown in the tables below.

To estimate the prior and likelihood information, the Monte Carlo method from the first appendix is used. The normalized likelihood is approximated as the posterior under the same prior with $\sigma^2 = 100$, larger than the variances used in the experiment. Minimum classification error is achieved for a small value of prior information (1.22), but grows as prior information increases and decreases. 

The second experiment uses the prostate cancer regression dataset from the lasso2 R package \cite{lasso2}, which has a continuous outcome of interest (log-cancer volume) and 8 predictors. There are 97 data points in this data set, of which 75 are randomly chosen for training, and the remaining are chosen for the test set. The model used is:
\begin{itemize}
\item $\beta \sim MVN(0,\sigma^2I)$
\item $Y_i|\beta,X_i \sim N([X_i^{T}\beta], 1)$, where $\beta \in \mathbb{R}^{9}$, $X_{i} \in \mathbb{R}^9$ are the background covariates  (and 1 for an offset term) for individual $i$ in the study, and $Y_i \in \mathbb{R}$ is the log-cancer volume for individual $i$.
\item $Y_i$ are independent conditional on $\beta$.
\end{itemize}
For all $i \in 1...97$.
 
As in the previous experiment, 100 samples of the posterior distribution of model coefficients are drawn for the test individuals in the study. The posterior predictive mean is taken as the final prediction, which is justifiable from a decision theoretic standpoint under a sum of squared error loss function, and mean square error (MSE) is adopted as the measure of predictive accuracy. The Monte Carlo method from the first appendix is applied to estimate prior and likelihood information. The normalized likelihood is approximated as the posterior under the same prior with $\sigma^2 = 100$.  A similar phenomenon is exhibited in that predictive error is minimized for a small value of the prior information. However, some of the estimates of prior information end up negative, and since the KL-divergence is non-negative, 0 is a better estimate for these cases.

The tables below delineate the results of these experiments. In either case, test error is smallest for a small value of prior information. While follow up studies and more repetitions are needed, this suggests that tuning hyper-parameters to allow for small prior information is a reasonable way to construct a Bayesian model that has good predictive performance. This could be useful if one would like to use an entire data set for the purpose of fitting a Bayesian model, as opposed to splitting it up for training, validation, and testing.

\begin {table}
\begin{center}
\begin{tabular}{ | l | l | l | l | }
\hline
$\sigma^2$ & Estimated Prior Inf. & Estimated Likelihood Inf.  & Error\\ \hline
 .1 & 78.9 & 39.5 & .267\\ \hline
 .5 & 11.4 & 40.8 & .244 \\ \hline
 1 & 2.29 & 39.1 & .229\\ \hline 
 1.25 & 1.22 & 55.0 & .217\\ \hline 
 1.5 & 1.95 & 67.0 & .252\\ \hline
 1.75 & 1.51 & 45.5 & .244 \\ \hline
 2 & .528 & 64.8 & .248 \\ \hline
 2.25 & 1.19 & 34.9 &  .236 \\ \hline
 2.5 & .666 & 63.7 & .236 \\ \hline
 2.75 & .143 & 75.1 & .221 \\ \hline
 3.00 & .948 & 47.7 & .236 \\ \hline
 \hline
  \end{tabular}
\caption{Results of Diabetes Prediction Experiment: Displayed are the value for $\sigma^2$ chosen,  the corresponding estimated prior and likelihood information for this choice of $\sigma^2$, and the classification error.}
\end{center}
\end {table}

\begin {table}
\begin{center}
\begin{tabular}{ | l | l | l | l | }
\hline
$\sigma^2$ & Estimated Prior Inf.& Estimated Likelihood Inf.  &  MSE\\ \hline
 .00001 &  Inf & 36.6 & 1.00\\ \hline
 .0001 & 378 & 3.43 & .982 \\ \hline
 .001 & 128 & 9.11 & .833\\ \hline 
 .01 & -2.71 (0) & 25.1 & .647 \\ \hline 
 .1 & .350 & 29.0 & .591\\ \hline
 1 & -.061 (0) & 337 & .603 \\ \hline
 10 & -.006 (0) & Inf & .813 \\ \hline
 \hline
\end {tabular}
\caption {Results of Log-Prostate Cancer Prediction Experiment: Displayed are the value for $\sigma^2$ chosen, the corresponding estimated prior and likelihood information for this choice of $\sigma^2$, and the mean square error.}
\end{center}
\end {table}

\section{Conclusion}
The two metrics constructed appear to be reasonable measures of prior and likelihood information, as is evidenced by their theoretical properties, analytical tractability in basic conjugate models, and use in applied contexts. 
Additional lines of investigation may stem from applying these measures to more complicated models and utilizing other methods for computing prior and likelihood information.


\section*{Acknowledgment}
Mentors and anonymous reviewers are thanked for guidance and feedback.

\section{Appendix}
\subsection{A Monte Carlo Method for Estimating Prior and Likelihood Information}
Here Monte Carlo methods to estimate the prior and likelihood information are developed, and the variances of the resultant estimators are quantified with the delta method. Without essential loss of generality, these methods are developed to estimate the prior information. First note the following identities:
\begin{eqnarray*}
D_{KL}(p(\theta|y_{obs}), L_{\theta}(\theta)) &=& E_{\theta|y_{obs}}[Log(\frac{p(\theta|y_{obs})}{L_{\theta}(\theta)})] \\
                                            &=& E_{\theta|y_{obs}}[Log(\frac{c_1p(y_{obs}|\theta)p(\theta)}{c_2p(y_{obs}|\theta)})] \\
                                            &=& Log(c_1/c_2)+ E_{\theta|y_{obs}}[Log(p(\theta))]
\end{eqnarray*}
Where $c_1$ and $c_2$ are the normalizing constants for the posterior and likelihood respectively. Assume i.i.d samples $\theta_1,...\theta_{N}$ from the posterior. Then by the identity 
\begin{eqnarray*}
 E_{\theta|y_{obs}}[\frac{1}{p(\theta)}] &=& c_1/c_2 
\end{eqnarray*}
a natural Monte Carlo estimator for the prior information is:
\begin{eqnarray*}
Log[(N_1)^{-1}\sum_{i=1}^{N_1} p(\theta_i)^{-1}]+(N-N_1)^{-1}\sum_{i=N_1+1}^{N}(Log(p(\theta_i))
\end{eqnarray*}
where $N_1$  of $N$ posterior samples have been chosen to estimate  $Log(c_1/c_2)$. By the WLLN and the continuous mapping theorem, this estimator converges in probability to the prior information, assuming that both $N_1$ and $N$ approach $\infty$ (so, for instance, $N_1$ could be some fraction of $N$). Since each component of the sum is consistent, the sum must also be consistent by basic results on convergence of sums in probability to the sum of their individual limits. However, since for finite samples it is possible for this estimator to be negative, yet the KL-divergence is non-negative, it is suggested to set negative values to 0 (e.g., see the second experiment of section four). The asymptotic variance of the estimator can be approximated using the delta method with the Log(.) transformation, yielding $(c_2/c_1)^2V_{\theta|y_{obs}}[1/p(\theta)] + V_{\theta|y_{obs}}[Log[p(\theta)]]$.

If one generates normalized-likelihood samples, a consistent estimator for $Log(c_1/c_2)$ is $Log[N_1/ \sum_{i=1}^{N_1} p(\theta_i)]$, where $\theta_i$ are draws from the normalized likelihood, whose asymptotic variance can be approximated with the delta method as $(c_1/c_2)^2V_{L(\theta)}[p(\theta)]$. This estimator is based on the identity:
\begin{eqnarray*}
 E_{L(\theta)}[{p(\theta)}] &=& c_2/c_1
\end{eqnarray*}
More efficient Monte Carlo estimators may be derived by using other methods for estimating the ratio of normalizing constants, such as in \cite{meng1996simulating}. Moreover, by the ergodic theorem for Markov chains, the convergence results presented in this section hold when the posterior and normalized likelihood samples are generated by a suitable Markov chain Monte Carlo algorithm. 

\subsection{Conditions for When Prior and Likelihood Information are Well Defined.}
Here are a set of sufficient (but not necessary) conditions for when the likelihood information is well defined, with the consequence that the likelihood information is bounded but possibly negative.  Precisely:

\textbf{Theorem A.1:}
Without essential loss of generality, assume the prior and posterior are continuous, the posterior is integrable, the prior and (unnormalized) likelihood are  bounded and the Shannon differential entropy of the posterior $H(\theta|y_{obs}) = -D_{KL}(\theta|y_{obs},1)$ exists. Then the likelihood information is bounded (and possibly negative).

\textbf{Proof:} Let $UB_{\theta}$ be an upper bound for $p(\theta)$ and $UB_{\l}$ be an upper bound for $p(y_{obs}|\theta)$.  An upper bound for the likelihood information is derived as follows:
\begin{eqnarray*}
v &=& \int_{\theta|y_{obs}} Log[\frac{p(\theta|y_{obs})}{p(\theta)}]p(\theta|y_{obs})d\theta \\
   &=& \int_{\theta|y_{obs}} Log[\frac{p(y_{obs}|\theta)p(\theta)}{p(y_{obs})p(\theta)}]p(\theta|y_{obs})d\theta \\
   &=& \int_{\theta|y_{obs}} Log[\frac{p(y_{obs}|\theta)}{p(y_{obs})}]p(\theta|y_{obs})d\theta 
\end{eqnarray*}
Which is bounded above by $Log[UB_l]-Log[p(y_{obs})]$.This exists because the normalizing constant, $p(y_{obs})$, exists since the posterior is assumed to be integrable. To derive a lower bound, since $Log[\frac{p(\theta|y_{obs})}{p(\theta)}] \geq Log[\frac{p(\theta|y_{obs})}{UB_{\theta}}]$, $v \geq D_{KL}(p(\theta|y_{obs}), UB_{\theta}) = -H(\theta|y_{obs})-Log(UB_{\theta})$. 

Additionally, if it is further assumed that the likelihood is integrable (i.e., the normalized likelihood is well-defined), then an analogous argument can be made to bound prior information from above.

\subsection{An Observation Connecting the Exponential Family (EF) of Distributions to the Natural Exponential Family (NEF) of Distributions}
Let 
\begin{eqnarray*}
f(y) &=& h(y)Exp[\eta^TT(y)-\psi(\eta)]
\end{eqnarray*}
be a distribution in the exponential family, where $\eta, T(y) \in \mathbb{R}^N$.
The likelihood function is this density as a function of the underlying parameters, that is 
\begin{eqnarray*}
L(\eta) = h(y)Exp[-\psi(\eta)]Exp[T(y)^T\eta]
\end{eqnarray*}
Assuming $\int_\mathbb{R^N} L(\eta)d\eta < \infty$, one can form a probability distribution $L*(\eta) \equiv L(\eta)/\int_\mathbb{R^N} L(\eta)d\eta$. Then $L*(\eta)$ is proportional to $Exp[-\psi(\eta)]Exp[T(y)^T\eta]$, which is in the NEF with parameter $T(y)$. 



\bibliographystyle{IEEEtran}
\bibliography{IEEEabrv,references_edit}
\end{document}